\documentclass[11pt,a4paper]{article}
\renewcommand\textsuperscript[1]{}

\newcommand{\hh}{\mathbf{h}}
\newcommand{\sss}{\mathbf{s}}
\newcommand{\cc}{\mathbf{c}}
\newcommand{\xx}{\boldsymbol{x}}
\newcommand{\yy}{\boldsymbol{y}}
\newcommand{\eos}{\textsc{eos}}

\usepackage{emnlp2017}
\usepackage{times}
\usepackage{url}
\usepackage{latexsym}
\usepackage{amsmath}
\usepackage[small]{caption}
\usepackage{arydshln}
\usepackage{booktabs}
\usepackage{tipa}

\usepackage{microtype}
\usepackage{adjustbox}
\usepackage{microtype}
\usepackage{inconsolata}

\hypersetup{
           breaklinks=true,   
           colorlinks=true,   
           pdfusetitle=true,  
        }

\usepackage{soul}
\newcommand{\Note}[1]{}
\definecolor{brilliantlavender}{rgb}{0.96, 0.73, 1.0}
\definecolor{thistle}{rgb}{0.847, 0.749, 0.847}

\newcommand{\saveforCR}[1]{}

\emnlpfinalcopy 
\def\emnlppaperid{127} 

\newcommand{\word}[1]{{\em #1}}

\usepackage{cleveref}
\crefname{section}{\S}{\S\S}
\Crefname{section}{\S}{\S\S}
\crefname{table}{Table}{}
\crefname{figure}{Figure}{}
\crefname{algorithm}{Algorithm}{}
\crefname{appendix}{Appendix}{}
\crefname{equation}{Equation}{}
\crefformat{section}{\S#2#1#3}  

\title{Paradigm Completion for Derivational Morphology}
\author{Ryan Cotterell\raise1.0ex\hbox{\normalsize \textnormal \textschwa}
\quad Ekaterina Vylomova\raise1.0ex\hbox{\textnormal {\normalsize \textipa{H}}}
\quad Huda Khayrallah\raise1.0ex\hbox{\normalsize \textnormal {\textschwa}} \\
\textbf{Christo Kirov}\raise1.0ex\hbox{\normalsize \textnormal 
 \textschwa}\quad \textbf{David Yarowsky}\raise1.0ex\hbox{\normalsize  \textnormal  \textschwa} \\
  {\raise1.0ex\hbox{\normalsize \textschwa}}Center for Language and Speech Processing, Johns Hopkins University, USA \\
  \raise1.0ex\hbox{\normalsize \textipa{H}}Department of Computing and Information Systems, University of Melbourne,  Australia \\
  {{ {\tt \href{mailto:ryan.cotterell@jhu.edu}{ryan.cotterell@jhu.edu}} \quad {\tt \href{mailto:vylomovae@unimelb.edu.au}{vylomovae@unimelb.edu.au}}}}
 }

\date{}

\begin{document}
\maketitle
\begin{abstract}
  The generation of complex derived word forms has been an overlooked
  problem in NLP; we fill this gap by applying neural
  sequence-to-sequence models to the task. We overview the theoretical
  motivation for a paradigmatic treatment of derivational morphology,
  and introduce the task of derivational paradigm completion as a
  parallel to inflectional paradigm completion. State-of-the-art
  neural models, adapted from the inflection task, are able to learn a
  range of derivation patterns, and outperform a non-neural baseline
  by 16.4\%. However, due to semantic,
  historical, and lexical considerations involved in derivational
  morphology, future work will be needed to achieve performance
  parity with inflection-generating systems.
\end{abstract}

\section{Introduction}
Unlike inflectional morphology, which produces grammatical variants of
the same core lexical item (e.g., {\em take$\mapsto$takes}),
derivational morphology is one of the key processes by which new
lemmata are
created. For example, the English verb {\em corrode} can evolve into
the noun {\em corrosion}, the adjective {\em corrodent}, and numerous
other complex derived forms such as \word{anticorrosive}. Derivational
morphology is often highly productive, leading to the ready creation
of neologisms such as \word{Rao-Blackwellize} and
\word{Rao-Blackwellization}, both originating from the Rao-Blackwell
theorem. Despite the prevalence of productive derivational morphology,
however, there has been little work on its generation. Commonly used derivational 
resources such as NomBank \cite{meyers-EtAl:2004:HLTNAACL} are still finite. Moreover, the
complex phonological and historical changes (e.g., the adjectivization
\word{corrode}$\mapsto$\word{corrosive}) and affix selection (e.g.,
choosing between English deverbal suffixes \word{-ment} and \word{-tion}) make
generation of derived forms an interesting and challenging problem for NLP.\looseness=-1


In this work, we show that viewing derivational morphological processes as
paradigmatic may be fruitful for generation. This means that there are
a number of well-defined form-function pairs associated with a core
word. For example, a typical English verb may have five forms in its
inflectional paradigm, corresponding to its base (\word{take}),
past tense (\word{took}), past participle (\word{taken}), progressive (\word{taking}) and
third-person singular (\word{takes}) forms. These forms are related by
a consistent set of relations, such as affixation. Similarly, a verb
may have several slots in its derivational paradigm: The form \word{take}
has the agentive nominalization \word{taker}, and the abilitative
adjectivization \word{takable}. Note there are also consistent patterns
associated with each derivational slot, e.g., the \word{-er} suffix
regularly produces the agentive.

Exploiting this paradigmatic characterization of derivational
morphology allows us to create a statistical model capable of
generating derivationally complex forms. We apply state-of-the-art
models for inflection generation, which learn mappings from fixed
paradigm slots to derived forms. Empirically, we compare results for
two models on the new task of derivational paradigm completion: a
neural sequence-to-sequence model and a standard non-neural
baseline. Our best neural model for derivation achieves 71.7\%
accuracy, beating the non-neural baseline by 16.4
points. Nevertheless, we note this is about 25 points lower than the
equivalent model on the English inflection task (and even 20 points lower than
the model's performance on the harder Finnish inflection generation).  These results point to additional complications in derivation that require
more elaborate models or data annotation to overcome. While inflection
generation is becoming a solved problem \cite{cotterell-conll-sigmorphon2017}, derivation generation is still
very much open.

\section{Derivational Morphology}\label{sec:derivational}

\begin{table}
  \centering
  \begin{adjustbox}{width=1.\columnwidth}
  \begin{tabular}{l  l l } \toprule
    { Semantics} & { POS} & { Affix} \\ \cmidrule(r){1-1} \cmidrule(r){2-3}
    {\sc negation} & {\tt J}$\rightarrow${\tt J} & \word{un-}, \word{in-}, \word{il-}, \word{ir-}  \\ \midrule
    {\sc origin} & {\tt N}$\rightarrow${\tt J} & \word{-an}, \word{-ian}, \word{-ish}, \word{-ese} \\
    {\sc relation} & {\tt N}$\rightarrow${\tt J} & \word{-ous}, \word{-ious}, \word{-eous} \\
    {\sc diminutive} & {\tt N}$\rightarrow${\tt N} & \word{-ette} \\ \midrule
    {\sc repeat} & {\tt V}$\rightarrow${\tt V} & \word{re-} \\
    {\sc patient}  & {\tt V}$\rightarrow${\tt N} & \word{-ee} \\
    {\sc result} & {\tt V}$\rightarrow${\tt N} & \word{-ment}, \word{-ion}, \word{-tion}, \word{-tion}, \word{-al}, \word{-ure} \\
    {\sc agent} & {\tt V}$\rightarrow${\tt N} & \word{-er}, \word{-or}, \word{-ant}, \word{-ee} \\
    {\sc potential} & {\tt V}$\rightarrow${\tt J} & \word{-able},\word{-abil}, \word{ible} \\ \bottomrule
  \end{tabular} 
  \end{adjustbox}
  \caption{A partial list of derivational transformations in English with corresponding POS changes and semantic labels.}
  \label{tab:deriv-sem}
  \setlength\belowcaptionskip{-15pt}
\end{table}


The generation of derived forms is structurally similar to the
generation of inflectional variants, but presents additional
challenges for NLP. Here, we provide linguistic background comparing
the two types of morphological processes. \saveforCR{Historically,
  neither structuralist \cite{harris1981morpheme} nor generative
  linguistics linguistics \cite{chomsky2014aspects} made any clear
  distinction between inflection and
  derivation. \newcite{bloomfield1933language} noted that there was no
  obvious criteria for separating the two. \newcite{aronoff1976word}
  was one of the first to try distinguishing characteristics.}

\paragraph{Inflection and Derivation.}
Inflectional morphology primarily marks semantic features that are
necessary for syntax, e.g., gender, tense and aspect. Thus, it follows
that in most languages\saveforCR{\footnote{There are some borderline
    cases in Australian Aboriginal languages}} inflection never
changes the part of speech of the word and often does not change its
basic meaning.  The set of inflectional forms for a given lexeme is
said to form a paradigm, e.g., the full paradigm for the verb \word{to
  take} is $\langle$\word{take}, \word{taking}, \word{takes},
\word{took}, \word{taken}$\rangle$. Each entry in an inflectional paradigm is
termed a slot and is indexed by a syntacto-semantic category, e.g.,
the {\sc past} form of \word{take} is \word{took}. We may
reasonably expect that all English verbs---including neologisms---have
these five forms.\footnote{Only a handful of English irregulars
  distinguish between the past tense and the past participle, e.g.,
  \word{took} and \word{taken}, and thus have five \emph{unique} forms in their
  verbal paradigms; most English verbs have four \emph{unique} forms.} Furthermore, there is typically a fairly regular
relationship between a paradigm slot and its form (e.g., add \word{-s}
 for the third person singular form).
Derivational morphology, on the other hand, often changes the core
part of speech of a word and makes more radical changes in meaning. In
fact, derivational processes are often subcategorized by the
part-of-speech change they engender, e.g.,
\word{corrode}$\mapsto$\word{corrosion} is a deverbal nominalization. 
 
\begin{table*}
  \centering
    {\em
  \begin{tabular}{l llll} \toprule
    \textnormal{Base} & \textnormal{-er/-or} & \textnormal{-ee} & \textnormal{-ment/-tion} & \textnormal{-able/-ible} \\  \hdashline
    \textnormal{POS} & {\tt V}$\mapsto${\tt N} & {\tt V}$\mapsto${\tt N} & {\tt V}$\mapsto${\tt N} & {\tt V}$\mapsto${\tt J}  \\
    \textnormal{Semantic} & {\sc agent} & {\sc patient} & {\sc result}  & {\sc potential} \\  \cmidrule(r){1-1} \cmidrule(r){2-5}
    animate    & animator    & ---        & animation                 & animatable \\
    attract    & attractor   & attractee  & attraction          & attractable \\
    ---        & aggressor   & aggressee  & aggression         & ---          \\
    employ     & employer    & employee   & employment                & employable \\ 
    place      & placer      & ---        & placement                 & placeable \\
    repel      & repeller    & repelee    & repellence           & repellable \\
    escape     & escapee     & ---        & ---                     & escapable \\
    corrode    & corroder    & ---        & corrosion            & corrosible \\
    derive     & deriver     & derivee    & derivation         & derivable \\
    \bottomrule
  \end{tabular}
    }
  \caption{Partial derivational paradigm for several English verbs; semantic gaps are indicated with ---. }
  \label{tab:derivational-paradigms}
  \vspace{-10pt}
\end{table*}


\paragraph{Derivational Paradigms.}
Much like inflection, derivational processes may be organized into
paradigms, with slots corresponding to more abstract lexico-semantic
categories for an associated part of speech
\cite{corbin1987morphologie,booij2008paradigmatic,stekauer}.
\newcite{lieber2004morphology} presents one of the first theoretical
frameworks to enumerate a set of derivational paradigm slots,
motivated by previous studies of semantic primitives by
\newcite{wierzbicka1988semantics}. A partial listing of
possible derivational paradigm slots for base English adjectives,
nouns, and verbs is given in \cref{tab:deriv-sem}.  The list contains
several productive cases. A key difficulty comes
from the the fact that the mapping between semantics and suffixes is
not always clean; \newcite{lieber2004morphology} points out the
category {\sc agent} could be expressed by the suffix \word{-er} (as
in \word{runner}) or by \word{-ee} (as in \word{escapee}). However,
both \word{-er} and \word{-ee} may have the {\sc patient} role;
consider 
\word{burner} (``a cheap phone intended to be disposed of, i.e. burned'') and
\word{employee} (``one being employed''), respectively. We flesh out partial derivational
paradigms for several English verbs in
\cref{tab:derivational-paradigms}.

Unlike in inflectional paradigms, where we expect most cells to be
filled for any given base form, derivational paradigms  often
contain base-slot combinations that are not semantically
compatible, leading to the gaps in \cref{tab:derivational-paradigms}.\footnote{For instance, if suffix \word{-ee} marks a {\sc patient}
it is semantically not compatible with intransitive verbs, i.e.,
\word{$^*$sneezee} cannot be derived from intransitive
\word{sneeze}.} We also observe increased paradigm irregularity due to some derived forms becoming lexicalized at different points in history, differences in the language from which
the base word entered the target language (e.g., English roots of
Germanic and Latinate origin behave differently
\cite{bauer1983english}), as well as other factors that are not
obvious from the characters in
the base word (e.g., gender or number of the resulting noun).\looseness=-1


As an example of how difficult these factors can make derivation,
consider the wide variety of potential nominalizations corresponding
to the {\sc result} of a verb, e.g., \word{-ion}, \word{-al} and
\word{-ment}, \cite{jackendoff1975morphological}. While any particular
English verb will almost exclusively employ exactly one of these
suffixes (e.g., we have \word{refuse}$\mapsto$\word{refusal} and other
candidates \word{$^*$refusion} and \word{$^*$refusement} are
illicit),\footnote{Note some forms appear to have multiple
  nominalizations, e.g.,
  \word{deport}$\mapsto$\{\word{deportation},\word{deportment}\}, but
  closer inspection shows there is one regular semantic
  transformation per word sense: \word{deportation} is eviction, but \word{deportment} is behavior. } the information required to
choose the correct suffix may be both arbitrary or not easily
available.

\paragraph{Productivity.}
There is a general agreement in linguistics that frequently used
complex words become part of the lexicon as wholes, while most others are likely to
be constructed  from constituents \cite{bauer2001morphological,aronoff2014productivity}; the latter ones typically follow
derivational patterns, or rules, such as adding \word{-able} to
express potential or ability or applying \word{-ly} to convert
adjectives into adverbs. These patterns typically present two
essential properties: productivity and restrictedness. Productivity
relates to the ability of a pattern to be applied to any novel base form to create a
new word, potentially on-the-fly.
One example of such a productive transformation is adding \word{-less} (privative construction), which may attach to almost any noun to form 
an adjective. Moreover, the resulting form's meaning is compositional and predictable. Many derivational suffixes in English are of this type. On the other 
hand, some patterns are subject to semantic, pragmatic, morphological or phonological restrictions. 
Consider the English patient suffix \word{-ee}, which cannot be
attached to a base ending in /i(:)/, e.g., it cannot be attached to the
verb \word{free} to form
\word{freeee}. Restrictedness is closely related to productivity, i.e., highly productive rules are less restricted. A parsimonious model of derivational morphology would describe forms using productive rules when possible, but may store forms with highly restricted patterns directly as full lexical items.\looseness=-1

\paragraph{A Note On Terminology.}
We would like to make a subtle, but important point regarding
terminology: the phrase {\em morphologically rich} in the NLP
community almost exclusively refers to {\em inflectional}, rather than
{\em derivational} morphology. For example, English is labeled as
morphologically impoverished, whereas German and Russian are
considered morphologically rich, e.g., see the introduction of
\newcite{tsarfaty2010statistical}. As regards derivation, English is
quite complex and even similar in richness to German or Russian
as it contains productive formations from two substrata: Germanic and
Latinate. From this perspective, English is
very much a morphologically rich language. Indeed, a corpus study on
the Brown Corpus showed that the {\em majority} of English words are
morphologically complex when derivation is considered
\cite{morphological-cues}.
Note that there are many languages that exhibit neither rich
inflection nor rich derivational morphology, e.g., Chinese, which
most commonly employs compounding for word formation
\cite{sino-tibetan}.\looseness=-1

%


\section{Task and Models}
We discuss our two systems for derivational paradigm completion
and the results they achieve.

\subsection{Data}\label{sec:data}
We experiment on English derivational triples extracted from NomBank
\cite{meyers-EtAl:2004:HLTNAACL}.\footnote{There are few resources annotated for derivation
  in non-English languages, making wider experimentation difficult.}
Each triple consists of a base form, the semantics of the derivation
and a corresponding derived form e.g., $\langle$\word{ameliorate},
{\sc result}, \word{amelioration}$\rangle$. Note that in this task we do not predict
whether a slot exists, merely what form it would take given the base and the slot. In terms of current
study, we consider the following derivational types: verb nominalization
such as {\sc result}, {\sc agent} and {\sc patient}, adverbalization
and adjective-noun transformations. We intentionally avoid
  zero-derivations. We also exclude overly orthographically distant pairs
  by filtering out those for which the Levenshtein distance exceeds
  half the sum of their lengths, which appear to be misannotations in
  NomBank. The final dataset includes 6,029 derivational samples,
  which we split into train (70\%), development (15\%), and test (15\%).\footnote{The dataset is available at {\small \url{http://github.com/ryancotterell/derviational-paradigms}}.}
  We also note that NomBank
annotations are often semantically more coarse-grained.

\begin{table}
  \centering
  \begin{adjustbox}{width=1.\columnwidth}
  \begin{tabular}{l ll ll l} \toprule
    & && \multicolumn{2}{c}{{\em 1-best}} & {\em 10-best }  \\
    & \multicolumn{2}{c}{{\em baseline}} & \multicolumn{2}{c}{{\em seq2seq}} &\multicolumn{1}{c}{{\em seq2seq}} \\ \cmidrule(r){2-3} \cmidrule(r){4-5} \cmidrule(r){6-6}
    & acc & edit & acc & edit & \multicolumn{1}{c}{{\em acc}}\\ \midrule
    all                                         & 55.3\% & 2.01        & 71.7\% & 0.97 & 84.5\%  \\ \midrule
    {\sc nominal} ({\tt J}$\mapsto${\tt N})   & 23.1\% & 3.45        & 35.1\% & 2.67 & 70.2\%  \\
    {\sc result} ({\tt V}$\mapsto${\tt N})  & 40.0\% & 2.24        & 52.9\% & 1.86 & 72.6\%\\
    {\sc agent} ({\tt V}$\mapsto${\tt N}) & 52.2\% & 0.94        & 65.6\% & 0.78 & 82.2\%\\
    {\sc adverb} ({\tt J}$\mapsto${\tt R})   & 90.0\% & 0.21        & 93.3\% & 0.18 & 96.5\%\\ \bottomrule
  \end{tabular}
  \end{adjustbox}
  \caption{Results under two metrics (accuracy and Levenshtein distance) comparing the non-neural baseline from the 201 SIGMORPHON shared task and the neural sequence-to-sequence model (both for 1-best and 10-best output).}
  \label{tab:results}
\end{table}
\subsection{Evaluation Metrics}
We evaluate on 3 metrics: accuracy, average edit distance, and
$F_1$. Accuracy measures how often system output exactly matches the gold
string. Edit distance, by comparison, measures the Levenshtein
distance between system output and the gold string. Finally, we calculate affix $F_1$
scores for individual derivational affixes. E.g., for \word{-ment}
precision is the number of words where the model correctly predicted \word{-ment}
(out of total predictions) and recall is the number
of words where the model correctly predicted out of the number of
true words.


\subsection{Baseline Transducer}

We train a simple transducer for each base-to-paradigm slot mapping in
the training set, identical to the baseline described in
\newcite{cotterell-EtAl:2016:SIGMORPHON}. This uses an averaged
perceptron classifier to greedily apply an output transformation
(substitution, deletion, or insertion) to each input character given
the surrounding characters and previous decisions.

\subsection{RNN Encoder-Decoder}
Following \newcite{kann-schutze:2016} on the morphological inflection task, we use an encoder-decoder gated
recurrent neural network \cite{bahdanau2014neural}. First, an encoder network encodes a
sequence: the concatenation of the characters of the input word
and a tag describing the desired transformation---both represented by
embeddings.  
This encoder is bidirectional and consists of two gated
RNNs \cite{cho2014properties}, one encoding the input in the forward
direction and one encoding in the backward direction. 
The output of
the two RNNs is the resulting hidden vectors $\overrightarrow{\hh_{i}}$
and $\overleftarrow{\hh_{i}}$. The hidden state is a concatenation
of the forward and backward hidden vectors, i.e., $\hh_i
=[{\overrightarrow{\hh_{i}}};{\overleftarrow{\hh_{i}}}]$.

The decoder also consists of an RNN, but is additionally equipped with
an attention mechanism.  The latter computes a weight for each of the
encoder hidden vectors for each character or subtag, which can be
roughly understood as giving a certain importance to each of the
inputs. 
The probability of the target sequence $\yy = (y_1, \ldots,
y_{|\yy|})$ given the input sequence $\xx = (x_1, \ldots, x_{|\xx|})$ is
modeled by
\begin{align}
\label{eq:2}
  p(\yy &\mid x_1,\ldots, x_{|\xx|}) \nonumber \\
  &= p(\eos \mid \yy)\prod^{|\yy|}_{t=1} p(y_t \mid y_1,
  \ldots, y_{t-1}, \cc_t) \nonumber
  \\ &= g(\eos,
  \sss_t, \cc_t)\prod^{|\yy|}_{t=1} g(y_{t-1},
  \sss_t, \cc_t),
\end{align}
where $\eos$ is a distinguished end-of-string symbol, $g$ is a multi-layer perceptron, $\sss_t$ is the hidden state of the
decoder and $\cc_t$ is the sum of the encoder states $\hh_{i}$, scored
by attention weights $\alpha_{i}(\sss_{t-1})$ that depend on the decoder
state: $\cc_t = \sum_{i=1}^{|\xx|} \alpha_{i}(\sss_{t-1}) \hh_{i}$.\looseness=-1

\paragraph{Input Encoding.}
We model this problem as a character translation problem, with special
encodings for the transformation tags that indicate the type of
derivation. For example, we treat the triple:
$\langle$\word{ameliorate}, {\sc result}, \word{amelioration}$\rangle$
as the source string {\footnotesize {\tt a m e l i o r a t e RESULT} }
and target string {\footnotesize {\tt a m e l i o
    r a t i o n}}.  This is similar to the encoding in
\newcite{kann-schutze:2016}.

\paragraph{Training.}
We use the Nematus
toolkit \cite{nematus}.\footnote{\url{https://github.com/rsennrich/nematus/}}
We exactly follow the recipe in \newcite{kann-schutze:2016}, the winning submission on the 2016 SIGMORPHON shared task for inflectional
morphology. Accordingly, we use a character embedding size of 300, 100
hidden units in both the encoder and decoder, Adadelta \cite{adadelta}
with a minibatch size of 20, and a beam size of 12. We train for 300 epochs
and select the test model based on the performance on the 
development set.



\section{Experimental Results}
\cref{tab:results} compares the accuracy of our baseline system with
the accuracy of our sequence-to-sequence neural network using the data
splits discussed in \cref{sec:data}. In all cases, the network
outperforms the baseline.  While 1-best performance is not nearly as
high as that expected from a state-of-the-art inflectional generation
system, the key point is that performance significantly increases when
considering the 10-best outputs. This suggests that the network is
indeed learning the correct set of possible nominalization
patterns. However, the information needed to correctly choose among
these patterns for a given input is not necessarily available to the
network. In particular, the network is only aware of important
disambiguating historical (e.g., is the input of Latin or Greek
origin) and lexical-semantic (e.g., is the input verb transitive or
intransitive) factors to the extent that they are implicitly encoded
in the input character sequence. We speculate that making these
additional pieces of information directly available as input features
will significantly improve 1-best accuracy.

Unfortunately, NomBank does not provide the necessary annotations in most cases.
For instance, there is no way to differentiate \word{actor}
and \word{actress} without gender.  It also does not distinguish the semantics of
some adjective nominalizations, e.g.,
\word{activism} and \word{activity}. Future work will
reannotate NomBank to make these finer-grained distinctions.

\paragraph{Error Analysis.} We observe mistakes on
less frequent suffixes, e.g., \word{-age}---we predict \word{$^*$draination} instead of \word{drainage}. 
Also, there are several cases where NomBank only lists one available form, e.g., \word{complexity},
and our model predicts \word{complexness}. We also see mistakes on irregular
adverbs, e.g., we generate \word{advancely} from \word{advance}, rather than \word{in-advance}, as well as in {\sc patient} nominalizations, e.g., the model produces \word{containee} in place of \word{content}---this last distinction is unpredictable.

\section{Related Work}
Previous work in unsupervised morphological {\em segmentation} and has
implicitly incorporated derivational morphology. Such systems attempt
to segment words into all constituent morphs, treating inflectional
and derivational affixes as equivalent. The popular Morfessor tool
\cite{creutz2007unsupervised} is one example of such an unsupervised
segmentation system, but many others exist, e.g.,
\newcite{poon-cherry-toutanova:2009:NAACLHLT09}, \newcite{TACL458}
        {\em inter alia}. Supervised segmentation and analysis models
        in the literature can also break down derivationally complex
        forms into their morphs, provided pre-segmented and labeled
        data is available for training
        \cite{ruokolainen-EtAl:2013:CoNLL-2013,cotterell-EtAl:2015:CoNLL,TACL1060}.
Our work, however, builds directly upon recent efforts in the {\em
  generation} of inflectional morphology \cite{DurrettDeNero2013,nicolai-cherry-kondrak:2015:NAACL-HLT,ahlberg-forsberg-hulden:2015:NAACL-HLT,rastogi-cotterell-eisner:2016:N16-1,faruqui:2016:infl}. We differ in that we focus on
derivational morphology.
In another recent line of work, \newcite{vylomova-EtAl:2017:EACLshort} predict
derivationally complex forms using sentential context. Our work differs
from their approach in that we attempt to generate derivational forms
divorced from the context, but the underlying neural sequence-to-sequence
architecture is quite similar.

\begin{table}
  \centering
  \begin{tabular}{ll ll ll} \toprule
    affix & $F_1$ & affix & $F_1$ & affix & $F_1$ \\ \cmidrule(r){1-2} \cmidrule(r){3-4} \cmidrule(r){5-6}
    -{\em ly} & 1.0 & -{\em ity} & 0.54 & -{\em ence} & 0.32\\
-{\em er} & 0.86 & -{\em ment} & 0.45 & -{\em ure} & 0.22 \\
-{\em ation} & 0.78 & -{\em ist} & 0.43 & -{\em ee} & 0.20 \\
-{\em or} & 0.59 & -{\em ness} & 0.40 & -{\em age} & 0.20 \\ \bottomrule
  \end{tabular}
  \caption{$F_1$ for various suffix attachments with the sequence-to-sequence model}
  \end{table}

\section{Conclusion}
We have presented a statistical model for the generation of
derivationally complex forms, a task that has gone essentially unexplored in
the literature. Viewing derivational morphology as paradigmatic, where
slots refer to semantic categories, e.g., \word{corrode}$+${\sc
  result}$\mapsto$\word{corrosion}, we draw upon recent advances
in the generation of inflectional morphology. Applying this method
works well, achieving an overall accuracy of 71.71\%, and beating a
non-neural baseline. Performance, however, is lower than on the task
of paradigm completion for inflectional morphology, indicating that
paradigm completion for derivational morphology is more challenging
than its inflectional counterpart.

\section*{Acknowledgements}
The authors would like to thank Jason Eisner and Colin Wilson for helpful discussions
about derivation.
The first author acknowledges funding from an NDSEG graduate fellowship.
This material is based upon work supported in part by the Defense Advanced Research Projects Agency (DARPA) under Contract No. HR0011-15-C-0113. Any opinions, findings and conclusions or recommendations expressed in this material are those of the authors and do not necessarily reflect the views of DARPA. 

\sloppy
\bibliography{derivational-paradigm-completion}
\bibliographystyle{emnlp_natbib}

\end{document}